\documentclass[11pt,a4paper]{article}
\usepackage[hyperref]{acl2020}
\usepackage{times}
\usepackage{latexsym}

\usepackage{microtype}

\definecolor{applegreen}{rgb}{0.55, 0.71, 0.0}

\usepackage{amsmath}
\usepackage{graphicx}
\usepackage{textcomp}
\usepackage{enumitem}
\usepackage{float}

\aclfinalcopy 

\title{Semantic Extractor-Paraphraser based Abstractive Summarization}

\author{Anubhav Jangra\thanks{* means equal contribution.} \\
  \normalsize{IIT Patna, India} \\
  \small{\texttt{anubhav0603@gmail.com}} \\\And
  Raghav Jain\footnotemark[1] \\
  \normalsize{DTU, India} \\
  \small{\texttt{raghavjain106@gmail.com}} \\\And
  Vaibhav Mavi\footnotemark[1] \\
  \normalsize{IIT Delhi, India} \\
  \small{\texttt{vaibhavg152@gmail.com}} \\\AND
  Sriparna Saha \\
  \normalsize{IIT Patna, India} \\
  \small{\texttt{sriparna.saha@gmail.com}} \\\And
  Pushpak Bhattacharyya \\
  \normalsize{IIT Bombay, India} \\
  \small{\texttt{pushpakbh@gmail.com}} \\
  }

\date{}

\begin{document}
\maketitle
\begin{abstract}
The anthology of spoken languages today is inundated with textual information, necessitating the development of automatic summarization models. In this manuscript, we propose an extractor-paraphraser based abstractive summarization system that exploits semantic overlap as opposed to its predecessors that focus more on syntactic information overlap. Our model outperforms the state-of-the-art baselines in terms of ROUGE, METEOR and word mover similarity (WMS), establishing the superiority of the proposed system via extensive ablation experiments. We have also challenged the summarization capabilities of the state of the art Pointer Generator Network (PGN), and through thorough experimentation, shown that PGN is more of a paraphraser, contrary to the prevailing notion of a summarizer; illustrating it's incapability to accumulate information across multiple sentences.

\end{abstract}


\section{Introduction} \label{sec:intro}

Over the past few years, the Internet has become the most convenient and preferred form of information sharing worldwide. The evolution of technology has made it possible for anyone to convey their knowledge, opinions and ideals to the world, resulting in an increasing surge of information hindering users from accessing desired content. This increasing need to obtain key information makes the task of summarization paramount. Text is the most widely adopted form of communication, be it for personal messaging\footnote{https://news.gallup.com/poll/179288/new-era-communication-americans.aspx} or for broadcasting, owing to its ability to convey almost any concept, its general flexibility to suit everyone's needs, and its less storage requirement (opposed to other modes of communication like audio and video). Text summarization is a problem at the very core of natural language processing, and has various applications in the spoken languages, including summarization of conversations, and public speeches.


Some works have been done in the field of reinforcement learning based text summarization \citep{dong2018banditsum,liu2018generative}, the most prominent architecture being extractor-abstractor (EXT-ABS) model \citep{chen2018fast}. Inspired from this architecture, in this manuscript, we have proposed an extractor-paraphraser system that uses semantic information overlap as the underlying guidance strategy. The model is further enhanced to surpass its limits using reinforcement learning, for which we have proposed a novel semantic overlap based reward function. Word Mover Similarity (WMS) \citep{clark2019sentence} is utilized to evaluate semantic similarity across generated sentences and the true ground truth summary sentences.

\begin{figure*}
    \centering
    \includegraphics[width=\textwidth, scale=1]{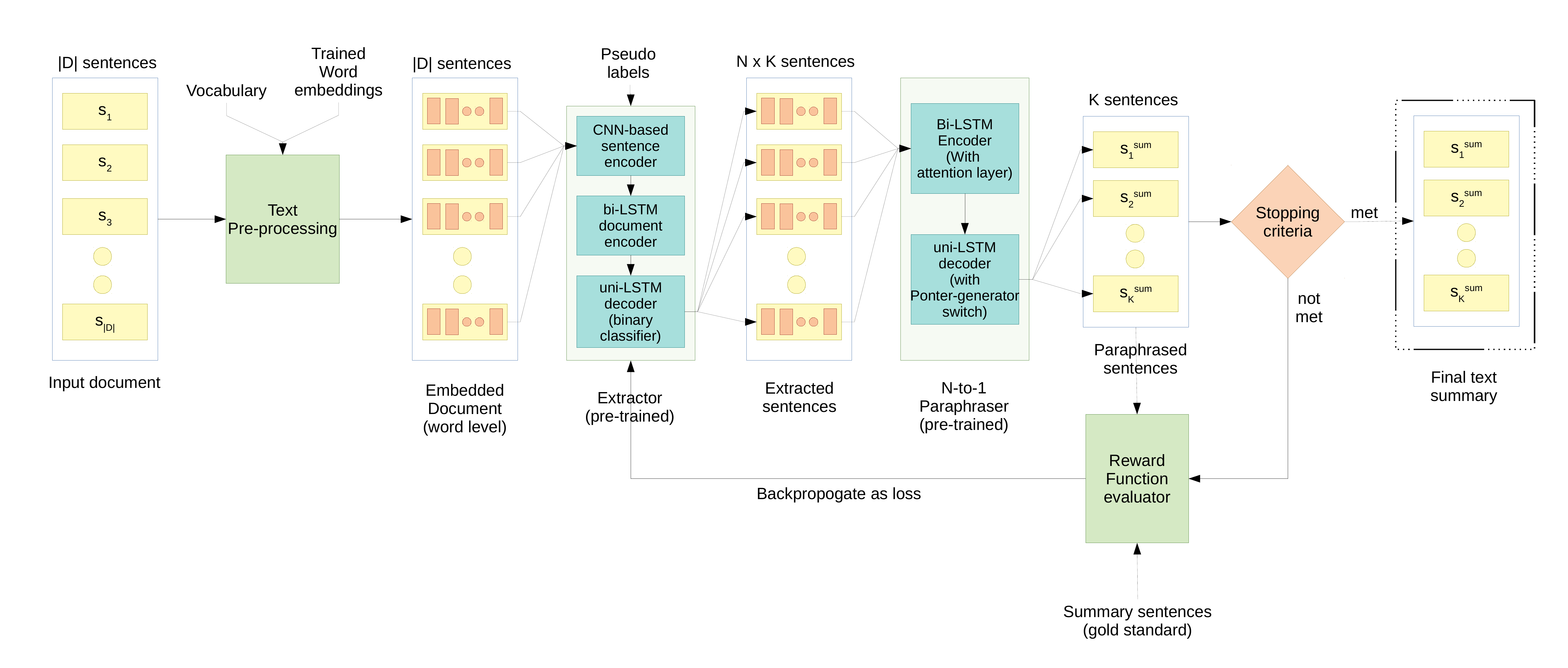}
    \caption{Proposed model architecture.}
    \vspace{-1em}
    \label{fig:arch}
\end{figure*}

We assume that paraphrasing is a relatively simpler task than abstractive summarization, with the underlying intuition that paraphrasing is a sub-problem within abstractive summarization. To bolster our hypothesis, experiments are conducted on the extractor-abstractor (EXT-ABS) model \cite{chen2018fast} and the Pointer Generator Network (PGN) \cite{DBLP:journals/corr/SeeLM17}, which is used as the basic abstraction unit in the former architecture. The results are rather staggering and reveal that 
the PGN model also paraphrases input document sentences, albeit implicitly.
The major contributions of the paper are as follows:
\begin{itemize}[leftmargin=*,noitemsep,nolistsep]
    \item A novel semantic overlap based reward function is proposed for reinforcement of extractor-paraphraser model.
    \item To the best of our knowledge, we are the first ever to discover the fact that PGN networks are indeed doing an implicit extraction-paraphrasing operation, revealing the true nature of existing abstractive summarization models.
\end{itemize}

The rest of this paper is structured as follows: In Section \ref{sec:rel_work} we have discussed related works of automatic text summarization. In Section \ref{sec:prop_meth} we have described the proposed model, and in Section \ref{sec:exp} we have stated the experimental setup and the datasets used. A thorough discussion and state the results  are provided in Section \ref{sec:res}, followed by the conclusion and future work in Section \ref{sec:conc}.

\section{Related Work} \label{sec:rel_work}
Automatic text summarization has been extensively researched over more than three decades, and has shown a lot of progress and promise over the course of time. Various approaches have been explored to tackle both extractive and abstractive summarization. Initial research \cite{paice1990constructing,kupiec1995trainable} focused on extractive summarization due to its easier setup. Various techniques ranging from integer linear programming \cite{galanis2012extractive}, graph based approaches \cite{mihalcea2004textrank,mihalcea2004graph}, genetic algorithms \cite{saini2019extractive2, saini2019extractive}, and neural networks \cite{nallapati2017summarunner,zhang2016multiview} have been adopted to solve the extractive summarization task. The majority of the research in abstractive summarization revolves around deep learning \cite{DBLP:journals/corr/SeeLM17,chopra2016abstractive,nallapati2016abstractive}. \citet{liu2018generative} proposed a generative adversarial network based model to generate document abstracts. A handful of works however also use ILP \cite{banerjee2015multi} and graph-based \cite{ganesan2010opinosis} techniques to attempt to solve the problem. A lot of domain specific summarization techniques have also been explored, like radiology findings summarization \cite{zhang2018learning}, across-time summarization \cite{duan2019across}, movie review summarization \cite{zhuang2006movie}, book summarization \cite{mihalcea2007explorations}, and customer review based opinion summarization \cite{pecar2018towards}. Lately, multi-modal summarization \cite{jangra2020text, jangra2020multi, zhu3multimodal, saini2020textual} has also gained popularity . 
\par
Recently, people have also explored reinforcement learning to tackle the problem of automatic text summarization in both extractive \cite{dong2018banditsum,gao2019reward} and abstractive domains \cite{xiao2020copy,chen2018fast}. \citet{chen2018fast} have proposed an extractor-abstractor architecture, separating the relevant data searching part and the paraphrasing part to individual modules. In this work, we have proposed a system inspired from \citet{chen2018fast}, stressing on the significance of semantic information over the traditional syntactic overlap. The literature on text summarization is rich, and has an abundance of survey papers \cite{yao2017recent,gambhir2017recent} to get an in depth overview of the domain.


\section{Proposed Method} \label{sec:prop_meth}


\textbf{Problem Definition}: Given the training data \{X, Y\} where $X = \{d_1, d_2, ... , d_N\}$ is the set of input documents and $Y = \{y_1, y_2, ... , y_N\}$ is the set of corresponding output summaries, the task of automatic summarization  is defined as the problem of discovering a function $f : X \mapsto Y$, such that $f(d_i) = y_i; \forall i \in \{1, 2, ... , N\}$. \par
We have proposed an extractor-paraphraser framework, which is inspired from the extractor-abstractor (EXT-ABS) framework introduced by \citet{chen2018fast}. The summarization function $f(.)$ is approximated as the composition, $f(d_i) = h(g(d_i))$, where the functions $g(.)$ and $h(.)$ are modeled as the \textit{extractor} and the \textit{paraphraser} components of the model, respectively. Given an input document $d_i = \{s^{d_i}_1, s^{d_i}_2, ... , s^{d_i}_{|d_i|}\}$, the extractor $g(.)$ extracts relevant sentences, acting as the primary noise filter. These extracted set of sentences are fed to the paraphraser $h(.)$, which accumulates the information into a concise gist of the extracted sentences, simulating the natural language generation module in the proposed system (Fig. \ref{fig:arch}).
The paraphraser in our system is capable of summarizing multiple extracted sentences to generate one sentence, in contrast with its predecessor `abstractor' from EXT-ABS model \cite{chen2018fast} which rephrases one extracted sentence at a time. The proposed framework consists of an `n-to-one paraphraser', that compiles $n$ sentences into one sentence, generating a richer summary. Formally, given a document $d_i = \{s^{d_i}_1, s^{d_i}_2, ... , s^{d_i}_{|d_i|}\}$, the extractor is defined as: 
\begin{equation} \label{eq:ext}
\begin{split}
    & g: X \mapsto Z^{|y_i|\times n}, \\
    & \text{ s.t. }
    g(d_i) = \{k_{j,l} \,|\, 1 \leq k_{j,l} \leq |d_i|\}_{l=1}^n {,}_{j=1}^{|y_i|} \\
\end{split}
\end{equation}

\noindent where $k_{p,q}$ represents the index of the $q^{th}$ extracted sentence corresponding to the $p^{th}$ sentence in the final gold summary. For modelling the same, we have used an encoder-decoder model, where the encoder consists of a temporal convolutional model \cite{kim2014convolutional} cascaded with a bidirectional LSTM network \cite{schuster1997bidirectional} and the decoder is a uni-directional LSTM model \cite{hochreiter1997long} (Fig. \ref{fig:arch}). \par 
\noindent The paraphraser function $h(.)$ is defined as follows:
\begin{equation} \label{eq:pph}
    h(\{k_{j,l}\}_{l=1}^n {,}_{j=1}^{|y_i|}, d_i) = \{p( \oplus_{l=1}^n s^{d_i}_{k_{j,l}}) \}_{j=1}^{|y_i|}
\end{equation}
where $\oplus$ represents the concatenation of sentences, $k_{j,l}$ represents the extractor output and $p$ is written as:
\begin{equation} \label{eq:pph2}
    p( s^{ext}_j ) = s^{y_i}_j 
\end{equation}
where $ s^{ext}_j = \oplus_{l=1}^n s^{d_i}_{k_{j,l}} $ and $ s^{y_i}_j$ is the $j^{th}$ gold summary sentence. As shown in Fig. \ref{fig:arch}, a pointer-generator framework \cite{DBLP:journals/corr/SeeLM17} is used to model the paraphraser. \par



\subsection{Training the Submodules}
In order to train the paraphraser beforehand, we need to mimic the output of an extractor\footnote{Note that the extractor's output can also be used to train the paraphraser at this step, however, since the extractor's weights get fine-tuned using reinforcement learning at a later stage, training paraphraser on extractor's non-ideal outputs might lead to unsatisfactory performance of the paraphraser.}. To fulfil this requirement, \textit{exemplary-extracted sentences} are generated using gold summaries.
Word-mover distance (WMD) \cite{kusner2015word} is adopted to create these exemplary-extracted sentences, with the motivation that it captures the semantic overlap between sentences better than its predecessors. For a training pair $\{d,y\}$, the labels are generated as:

\begin{equation} \label{eq:exemplary}
\begin{split}
    \forall l \in \{1,2,...,n\}: \\
    \forall s_j \in y: \\
    &k_{j,l} = argmin_i\{WMD(s^d_i, s^y_j)\}; \\
    &\forall s^d_i \in d - \{s^d_{k_{j^\prime,0}}\}_{j^\prime=1}^{j-1}
\end{split}
\end{equation}

where $k_{j,l}$ represents the $l^{th}$ exemplary- extracted sentence corresponding to the $j^{th}$ gold summary sentence, $s^y_j$.


The $n$ exemplary-extracted sentences corresponding to each summary sentence are concatenated and fed into the paraphraser as the input and the summary sentence is fed as the output. We argue that this should allow the paraphraser to generate information rich summaries. 
To facilitate the expected behaviour during the testing phase, we require the extractor to feed the paraphraser with input similar to the exemplary-extracted sentences. Hence, we first pre-train the extractor on these exemplary-extracted sentences as opposed to random initialization.
Cross-entropy loss is used for training the paraphraser and pre-training the extractor.

\subsection{Extractor agent}
\textbf{Enhancement using reinforcement learning:}
Here, the extractor is trained as part of an actor-critic model 
\cite{mnih2016asynchronous} which takes an action based on the current state and current value of parameters to maximize a given reward at each time step, where the action is to extract $n$ sentences, $\{k_{t,l}\}_{l=1}^n$, and the state refers to the set of document sentences, $d_i$, and already extracted sentences, $\{d_{k_{1, l}}, d_{k_{2, l}}, ... , d_{k_{t-1, l}}\}; \,l\in \{1, 2, ... , n\}$. The predicted sentences are concatenated and passed to the paraphraser to get an output sentence. A novel semantic-based reward function using word mover distance (WMD) is used as the reward function. Since the reward is to be maximised, WMD needs to be converted to a similarity function, for which, a generalised version of word mover similarity (WMS), proposed in \cite{clark2019sentence}, is used.
Formally, at a time step $t$, given an action ${j_{t,l}}$, and a summary sentence, $s_t$, the short term reward, $r$ is calculated as\footnote{Note that we obtain the word mover similarity proposed in \cite{clark2019sentence} for the case $a=0$, and $b=1$.}:
\begin{equation} \label{eq:reward}
r = \frac{a+1}{a+e^{b \times WMD( s_t^y, p(s_t^{ext}))}}
\end{equation}
where $s_t^{ext} = \oplus_{l=1}^n s^d_{k_{t,l}}$ , and $\oplus$ represents the concatenation of sentences, $\{s^d, s^y\}$ are the sentences belonging to a training pair and $a$ and $b$ are the hyper-parameters introduced\footnote{The hyperparameters are set to $a = 1$ and $b = 0.5$ after extensive experimentations, and are used throughout this paper.}.

To avoid redundant phrases and words, tri-gram avoidance through beam search \cite{paulus2017deep} is applied at a sentence level. For a fair comparison, details regarding the beam search reranking and other nuances of implementation are kept the same as \cite{chen2018fast}.

\begin{table*}[t!]
\centering
\caption{Evaluation scores for the generated text summary using ROUGE, METEOR and Word Mover Similarity (WMS). `Id-ext' refers to the ideal extractor experiments. The '-' denotes unavailability of a score.}
\label{tab:main}
\begin{tabular}{|l|ccccc|}
\hline
Models & ROUGE-1 & ROUGE-2 & ROUGE-L & METEOR & WMS \\
\hline
\multicolumn{6}{|c|}{\textbf{Extractive baselines}} \\
\hline
$EXT-ABS \,(Ext \,only)$ & 40.17 & 18.11 & 36.41 & \textbf{22.81} & - \\
$O2O \,(Ext \,only)$ & \textbf{40.97} & \textbf{18.42} & \textbf{37.35} & 21.86 & \textbf{14.28} \\
$M2O_{n=2} \,(Ext \,only)$ & 40.19 & 17.98 & 36.60 & 21.62 & 14.24 \\
\hline
\multicolumn{6}{|c|}{\textbf{Abstractive baselines}} \\
\hline
$PGN$ & 39.53 & \textbf{17.28} & 36.38 & 18.72 & 13.36 \\
$EXT-ABS \,(w/o \,RL)$ & 38.38 & 16.12 & 36.04 & \textbf{19.39} & - \\
$O2O \,(w/o \,RL)$ & \textbf{39.82} & 17.05 & \textbf{37.21} & 19.24 & \textbf{13.81} \\
$M2O_{n=2} \,(w/o \,RL)$ & 32.82 & 11.29 & 31.08 & 16.13 & 12.92 \\
\hline
\multicolumn{6}{|c|}{\textbf{Reinforced models}} \\
\hline
$EXT-ABS + RL_{ROUGE}$ & 40.88 & 17.8 & 38.53 & 20.38 & 13.7 \\
$O2O + RL_{ROUGE}$ & \textbf{41.32} & \textbf{18.16} & \textbf{38.89} & 20.52 & 14.56 \\
$EXT-ABS + RL_{WMS}$ & 40.82 & 17.82 & 38.45 & \textbf{21.47} & 14.42 \\
$O2O + RL_{WMS}$ & 41.2 & 18.12 & 38.81 & 21.34 & \textbf{14.6} \\
$M2O_{n=2} + RL_{ROUGE}$ & 39.71 & 16.7 & 37.32 & 18.25 & 13.52 \\
\hline
\multicolumn{6}{|c|}{\textbf{Ablation experiments (Ideal Extractor)}} \\
\hline
$Id-ext \,EXT-ABS $ & 49.73 & \textbf{26.53} & 47.2 & 24.36 & 19.34 \\
$Id-ext \,O2O$ & \textbf{50.04} & 26.31 & \textbf{47.34} & \textbf{24.61} & \textbf{20.14} \\
$Id-ext \,M2O_{n=2} $ & 47.00 & 23.37 & 44.23 & 22.22 & 17.99 \\
\hline
\end{tabular}%
\end{table*}

\section{Experiments} \label{sec:exp}

\subsection{Dataset}
For all the following experiments we have used the CNN / DailyMail dataset \cite{nallapati2016abstractive}, which contains online news articles, with the bullet highlights treated as the gold standard summaries. The experiments in this work are conducted on the non-anonymized version of this dataset. The dataset consists of 277,226 training, 13,368 validation and 11,490 test article-summary pairs. An article contains $\sim \!780$ tokens per document, whereas the summary consists of $\sim\!56$ tokens with the average number of sentences per summary being $\sim \!3.75$. An article sentence, on average contains $\sim \!30$ tokens.

\subsection{Comparative methods} 
To highlight the superiority of the proposed semantic-overlap based methodology over existing syntactic measures, we set the value of $n = 1$ for initial experiments. To compare the complexity of the summarization task with paraphrasing objectively, the experiments are further extended for $n = 2$ as well \footnote{We have limited our work to $n = 2$ since the Two-to-one baselines did not perform efficaciously. Experiments for $n > 2$ would be done in future works.}. We evaluate our model with sufficient baselines \footnote{Statistical analysis on all the variations of the proposed model has been done.}, including: \par
\noindent \textbf{Pointer-Generator network [$\mathbf{PGN}$]:} An encoder-decoder attention based framework for abstractive summarization \cite{DBLP:journals/corr/SeeLM17}. \par
\noindent \textbf{Pre-trained extractor - Extractor-Abstractor framework [$\mathbf{EXT-ABS \,(Ext \,only)}$]:} The pre-trained extractor of the extractor-abstractor framework proposed by \citet{chen2018fast} on ROUGE-L based exemplary-extracted sentences. \par
\noindent \textbf{Extractor-Abstractor framework without reinforcement [$\mathbf{EXT-ABS \,(w/o \,RL)}$]:} The extractor-abstractor framework proposed by \citet{chen2018fast} at a stage before reinforcing the extractor. \par

\begin{sloppypar}
\noindent \textbf{Extractor-Abstractor framework [$\mathbf{EXT-}$ $\mathbf{ABS + RL_{X}}$]:} The complete extractor-abstractor framework proposed \citet{chen2018fast} including X as the reward function for reinforcement learning (X is either ROUGE-L or the reward function defined in Eq. \ref{eq:reward}) along with beam search. \par
\end{sloppypar}

\noindent \textbf{Pre-trained extractor - One-to-one extractor-paraphraser model [$\mathbf{O2O \,(Ext \,only)}$]:} The pre-trained extractor of the proposed extractor-paraphraser framework (with $n=1$) on WMD based (Eq. \ref{eq:exemplary}) exemplary-extracted sentences. \par

\noindent \textbf{One-to-one extractor-paraphraser model without reinforcement [$\mathbf{O2O \,(w/o \,RL)}$]:} A particular setting of the proposed methodology where $n=1$, without any reinforcement or beam search. \par

\noindent \textbf{One-to-one extractor-paraphraser model [$\mathbf{O2O +}$ $\mathbf{RL_{X}}$]:} A particular setting of the proposed methodology where $n=1$, including extractor, paraphraser, and X as the reward function for reinforcement learning (X is either ROUGE-L or the reward function defined in Eq. \ref{eq:reward}) and beam search. \par

\noindent \textbf{Pre-trained extractor - Two-to-one extractor-paraphraser model [$\mathbf{M2O_{n=2} \,(Ext \,only)}$]:} The pre-trained extractor\footnote{For fair comparison of extraction capabilities across all models, we limit the model to output 4 sentences during evaluation on test dataset.} of the proposed extractor-paraphraser framework (with $n=2$) on WMS based (Eq. \ref{eq:exemplary}) exemplary-extracted sentences. \par

\noindent \textbf{Two-to-one extractor-paraphraser model without reinforcement [$\mathbf{M2O_{n=2} \,(w/o \,RL)}$]:} A specific case of the proposed many-to-one paraphrasing where $n = 2$. This specific model comprises of only the extractor and paraphraser modules. \par

\noindent \textbf{Two-to-one extractor-paraphraser model [$\mathbf{M2O_{n=2} + }$ $\mathbf{RL_X}$]:} A specific case of the proposed many-to-one paraphrasing where $n = 2$; including X as the reward function for reinforced extractor (X is either ROUGE-L or the reward function defined in Eq. \ref{eq:reward}) and beam search. \par

\begin{sloppypar}
\noindent \textbf{Ideal extractor-abstractor model [$\mathbf{Id-ext}$ $\mathbf{ \,EXT-ABS}$]:} 
To determine the performance of the abstractor component, the $EXT-ABS$ baseline is evaluated with the assumption that the extractor is ideal, subsequently feeding the exemplary-extracted sentences generated for the test data using ROUGE-L score to the paraphraser directly.
\end{sloppypar}

\begin{sloppypar}
\noindent \textbf{Ideal extractor-paraphraser model [$\mathbf{Id-ext}$ $\mathbf{ \,O2O}$]:}  Similar to the $Id-ext \,EXT-ABS$ setup, the 'n-to-1' paraphraser is evaluated with $n=1$, given that the extractor performs ideally (generating exemplary-extracted sentences as proposed in Eq. \ref{eq:exemplary}).  \par
\end{sloppypar}

\begin{sloppypar}
\noindent \textbf{Ideal extractor-paraphraser model [$\mathbf{Id-ext}$ $\mathbf{ \,M2O_{n=2}}$]:}  The 'n-to-1' paraphraser with $n=2$ and the exemplary-extracted sentences assumed as the extractor (generating exemplary-extracted sentences as proposed in Eq. \ref{eq:exemplary}). \par
\end{sloppypar}

\section{Results} \label{sec:res}
Results of different baselines and the proposed approach are discussed in this section. Table \ref{tab:main} illustrates that our proposed techniques perform better than the rest of the systems. We have used ROUGE-1, ROUGE-2, ROUGE-L \cite{lin-2004-rouge}, METEOR \cite{banerjee2005meteor} and word mover similarity (WMS) \cite{clark2019sentence} as evaluation metrics. We believe that ROUGE as an evaluation metric is incapable of judging the quality of an abstractive summary due to its emphasis on syntactic overlap over semantic overlap \cite{liu2016not,clark2019sentence,novikova2017we}. To overcome this, we have also used WMS as an evaluation metric. \par

\begin{figure}
    \centering
    \includegraphics[width=\linewidth, scale=0.5]{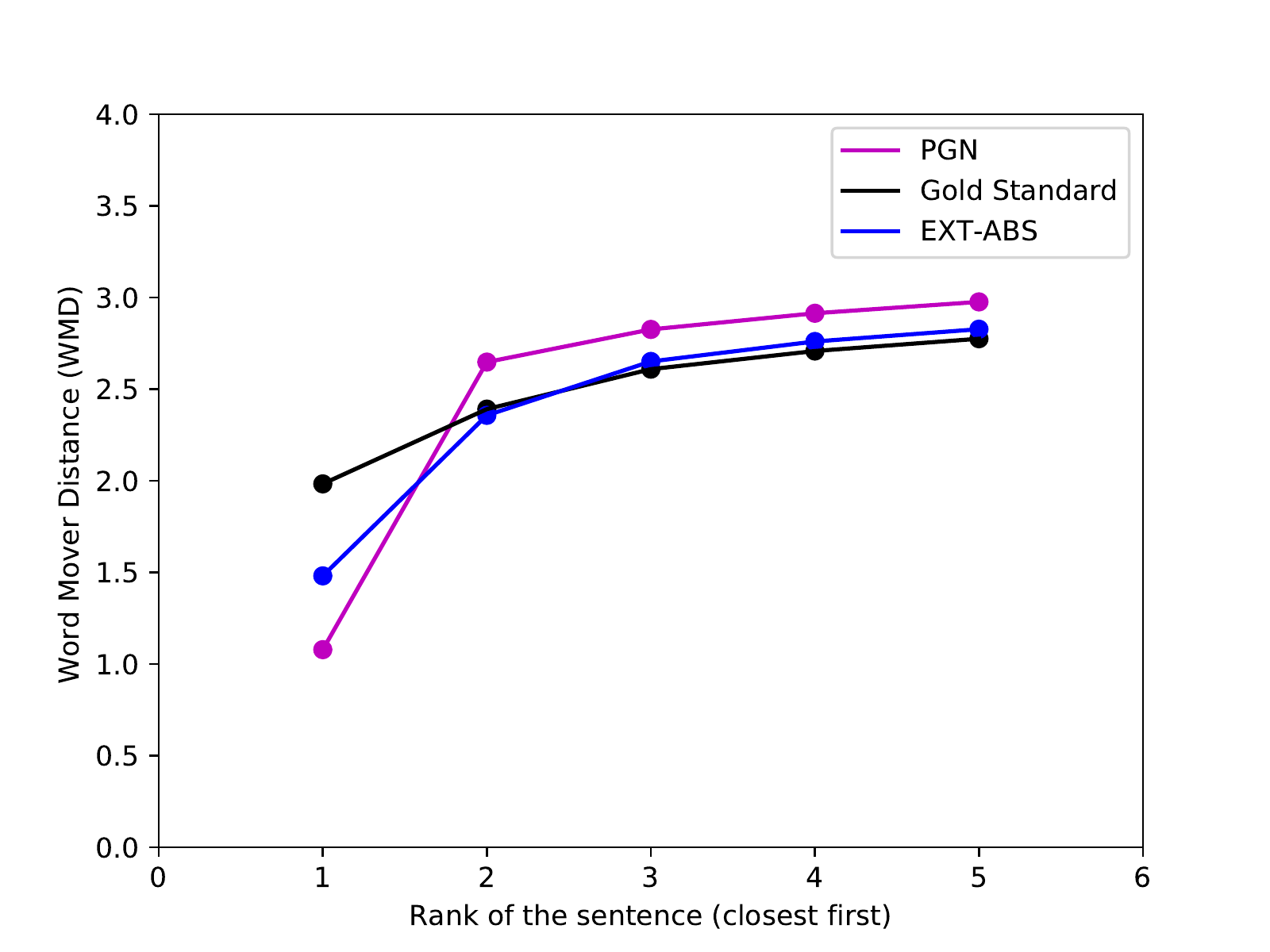}
    \vspace{-4mm}
    \caption{Average sentence distance (word mover distance) scores for most similar sentences.}
    \vspace{-0.5em}
    \label{fig:plot}
\end{figure}

\subsection{Semantic information overlap}
It is noticed that the proposed paraphraser in model $O2O \,$ $(w/o \,RL)$ outperforms the abstractor from $EXT-ABS \,$ $(w/o \,RL)$ in terms of ROUGE scores, while scoring marginally less in terms of METEOR. The extractor counterparts $EXT-ABS \,(Ext \, only)$ and $O2O \,(Ext \, only)$ also portray a similar tendency, with a wider gap in METEOR scores. We also observe that in the reinforced extractor models, the $O2O + RL_X$ setting surpasses the $EXT-ABS + RL_X$ setting in almost all the metrics\footnote{Here $X \in \{ROUGE, WMS\}$, which remains same when comparing the two models}. A similar trend is also observed in the ideal extractor experiments, where the $Id-ext \,O2O$  model beats the $Id-ext \,EXT-ABS$ model in WMS while keeping other evaluation scores comparable. The above mentioned observations illustrate the true capabilities of using semantic overlap based exemplary-extracted sentences.

Keeping the main model same as the $EXT-ABS$ framework, and changing the reward function from ROUGE-L to WMS (Eq. \ref{eq:reward}), it is observed that the latter ($EXT-ABS \,RL_{WMS}$) attains significantly better METEOR and WMS scores, while maintaining comparable ROUGE scores. However, the true capabilities of the WMS reward function (Eq. \ref{eq:reward}) come into play when we attach it with our extractor-paraphraser framework; $O2O \,RL_{WMS}$ bests every other models in terms of WMS, while keeping ROUGE and METEOR comparable with the best attained scores.

\begin{figure}[H]
    \centering
    \includegraphics[width=\linewidth, scale=0.5]{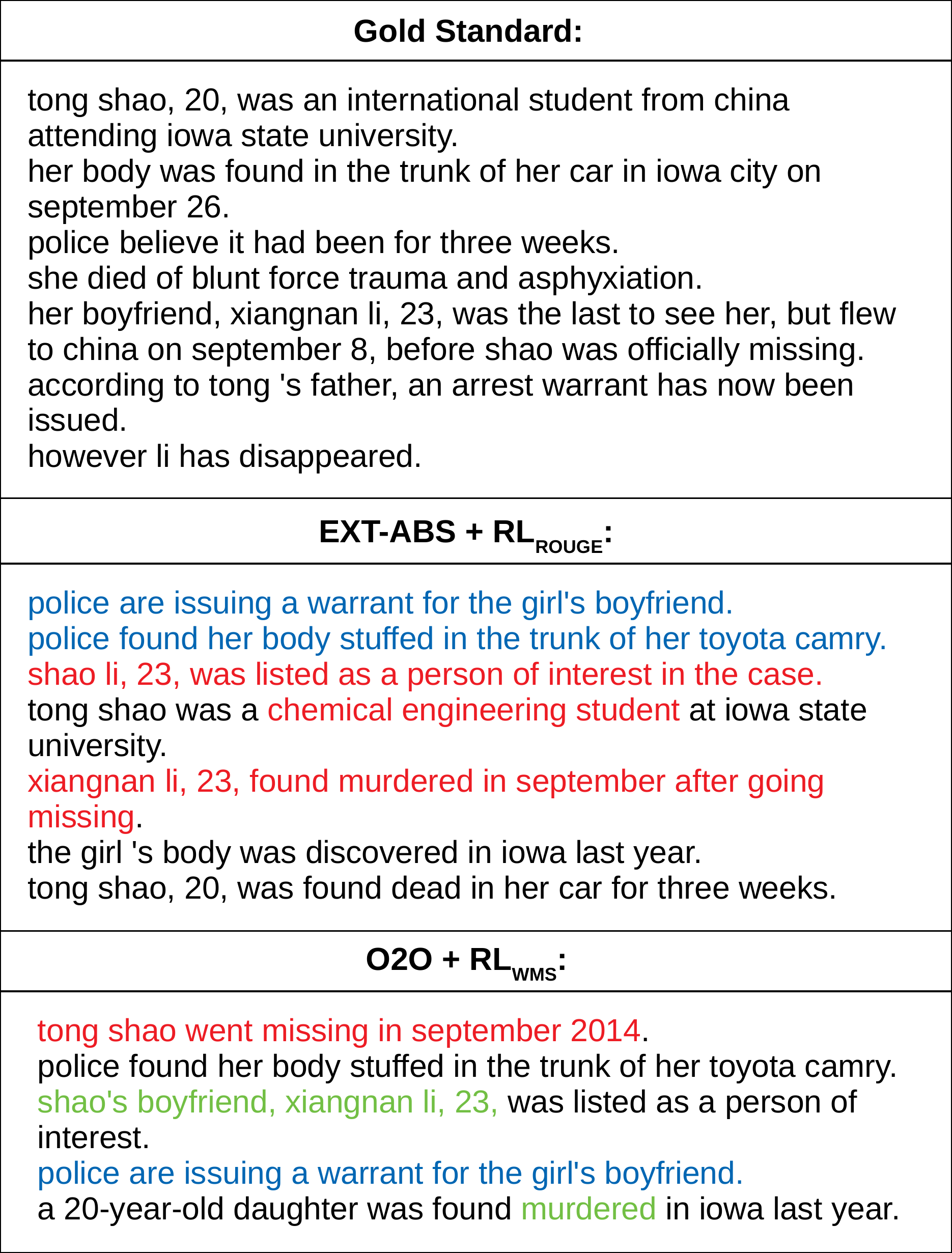}
    \vspace{-4mm}
    \caption{Example of generated summaries for $EXT-ABS + RL_{ROUGE}$ model, $O2O + RL_{WMS}$. The text in \textcolor{red}{red} denotes the novel information that is not present in the gold summary, the text in \textcolor{applegreen}{green} denotes the information overlap that is present exclusively in the generated summary and the text in \textcolor{blue}{blue} denotes the information covered in all three scenarios.}
    \vspace{-0.5em}
    \label{fig:summ}
\end{figure}

One critical observation is that the reward function introduces a bias in the evaluation process. It can be clearly observed from the fact that the model $EXT-ABS + RL_{ROUGE}$ obtains better ROUGE scores while it pales in comparison to $EXT-ABS + RL_{WMS}$ in terms of WMS. Since we stress that semantic information overlap is more significant than the syntactic overlap, we believe that WMS is better suited for the evaluation task as well as a better choice for the reward function. It is established by the fact that the models using WMS as the reward function attain comparable ROUGE scores as well (while the reverse is not true), indicating that incorporating semantic information can assist in capturing syntactic information as well. Examples of generated summaries for the $EXT-ABS \,RL_{ROUGE}$ and $O2O \,RL_{WMS}$ models are illustrated in Fig. \ref{fig:summ}. An important observation in the generated summaries is that the former model produces the name "shao li" which is not present in the input document or the gold standard summary, whereas this mistake is avoided by the $O2O \,RL_{WMS}$ model.

\subsection{Summarization vs paraphrasing}
Theoretically the $M2O_{n=2}$ setting should surpass the $O2O$ setting, since the former has extra input information at the paraphraser stage that the latter lacks. However, it is observed that this does not happen; in actuality, the $M2O_{n=2}$ model is outperformed by the $O2O$ model in all aspects (Table \ref{tab:main}). After manual scrutiny of \textit{exemplary-extracted sentence} pairs fed to the paraphraser and the generated sentences, it is observed that the expected accumulation of information does not take place. To quantify this observation, WMD based overlap of information is computed between the $exemplary-extracted \,sentence$ and the generated sentence. It is discovered that on average, the more similar sentence has a WMD of \textbf{1.775} while the other one obtains a value of \textbf{2.894}, illustrating the inability of the paraphraser to combine information across the two sentences into one. \par
Hence, we hypothesize that the PGN model intrinsically paraphrases one input sentence to generate the corresponding sentence in the generated summary, innately mimicking the extractor paraphraser behaviour.  
An experiment is formulated to evince the truth of this proposed hypothesis.
For this experiment, three different collections of documents are used: 1) ground truth summary, 2) summaries generated by PGN \cite{DBLP:journals/corr/SeeLM17}, and 3) summaries generated by the $EXT-ABS$ model \cite{chen2018fast}, all corresponding to the data in the test set. For every summary sentence, its semantic overlap (using WMD) is computed with every document sentence and distances of the closest $\alpha$ document sentences are reported in an increasing order. The results of this experiment can be seen in Fig. \ref{fig:plot}, and it is noticed that the average gap between similarities with respect to first and second most similar sentences is the highest in the case of the PGN model (approx. 1.57), while the value is around 0.41 for the gold standard summaries. The curve for gold standard has a steady slope, where as the slope for PGN rises quickly from the first point to the second point, and then follows a constant growth afterwards, illustrating a very high similarity with the most similar sentence, thus endorsing the hypothesis that the PGN intrinsically paraphrases one input sentence to generate one sentence in final summary.
\vspace{-1em}

\subsection{Error Analysis}
As the results illustrate in Table \ref{tab:main}, overall the one-to-one setting of the proposed n-to-one paraphraser performs better than the two-to-one setting, revealing a major blockade of the existing sequential frameworks - the incapability of combining multiple sentences into one, deviating from the ideal notion of `abstraction'. We also observe that the extractors outperform their corresponding $(w/o \,RL)$ counterparts, and are competing with the complete models as well to some extent, portraying the difficulty of abstractive summarization over sentence extraction, along with the inability of current evaluation metrics like ROUGE and METEOR to focus on semantic information over syntactic subtleties. As elucidated by the ideal extractor experiments, the paraphraser has great potential for even achieving state-of-the-art results in the summarization task, provided the extractor works ideally. Our future work would focus on mollifying the loss introduced at the extraction step, and obtaining a better harmony in the extractor-paraphraser network as a whole.


\section{Conclusion} \label{sec:conc}
In this paper, we propose the extractor-paraphraser model, that comprises of the n-to-one paraphraser and a novel word mover similarity based reward function. We show that the semantic overlap based techniques surpass the strong baselines that rely on syntactic information overlap. We also develop experiments to unveil that the state-of-the-art pointer-generator network (PGN) indeed paraphrases input sentences intrinsically, and is unable to merge two sentences into one agglomerate sentence when explicitly conditioned to do so,  changing our perception of sequence-to-sequence abstractive summarization models. We show that the existing works are still nowhere close to mimicking a true human summary, portraying the difficulty of true abstraction over its simplified formulation of extracting and then paraphrasing. \\

\vspace{1em}
\textbf{Acknowledgement:} Dr. Sriparna Saha would like to acknowledge the support of Early Career Research Award of Science and Engineering Research Board (SERB) of Department of Science and Technology, India to carry out this research.

\bibliography{refs}
\bibliographystyle{acl_natbib}

\end{document}